\documentclass[11pt]{article}
\usepackage[dvipdfm,colorlinks=true, citecolor=blue,linkcolor=red]{hyperref}
\usepackage{CJK}
\usepackage{acl2012}
\usepackage{times}
\usepackage{latexsym}
\usepackage{amsmath}
\usepackage{multirow}
\usepackage{url}

\usepackage{algorithm}
\usepackage{algorithmic}
\usepackage[dvips]{graphicx}

\DeclareMathOperator*{\argmax}{arg\,max}
\title{Reduce Meaningless Words for Joint Chinese Word Segmentation\\
and Part-of-speech Tagging}

\author{Kaixu Zhang \\
Xiamen University \\
{\tt kareyzhang@gmail.com} \\
  \And
Maosong Sun\\
Tsinghua University \\
{\tt sunmaosong@gmail.com} \\
}

\date{}

\begin{document}
\begin{CJK}{UTF8}{gbsn}
\maketitle
\begin{abstract}
Conventional statistics-based methods for joint Chinese word segmentation and part-of-speech tagging (S\&T) have generalization ability to recognize new words that do not appear in the training data. An undesirable side effect is that a number of meaningless words will be incorrectly created. We propose an effective and efficient framework for S\&T that introduces features to significantly reduce meaningless words generation. A general lexicon, Wikepedia and a large-scale raw corpus of 200 billion characters are used to generate word-based features for the wordhood. The word-lattice based framework consists of a character-based model and a word-based model in order to employ our word-based features.  Experiments on Penn Chinese treebank 5 show that this method has a 62.9\% reduction of meaningless word generation in comparison with the baseline. As a result, the F1 measure for segmentation is increased to 0.984.
\end{abstract}
\section{Introduction}
Chinese word segmentation is to segment Chinese sentence into words. Part-of-speech (POS) tagging is to further assign each segmented word a POS tag.

Statistics-based methods are used for joint Chinese word segmentation and part-of-speech tagging (S\&T). There are character-based methods \cite{ng_chinese_2004}, word-based methods \cite{zhang_joint_2008,zhang_fast_2010}, and hybrid methods employing both character-based features and word-based features \cite{kruengkrai_error-driven_2009}.

Conventional statistics-based methods for S\&T have the generalization ability to recognize new words that do not appear in the training data. However, as a side effect, they also erroneously create a number of meaningless words.

\begin{figure}[t]

\centering
\begin{tabular}{|rl|}
\hline
gold standard:& 拉脱维亚 (Latvija)\\
result:& 拉脱~~维亚\\
\hline
gold standard:& 射电~~望远镜 (radio telescope)\\
result:& 射~~电望~~远镜\\
\hline
\end{tabular}
\caption{There are errors that contain incorrectly created meaningless words such as ``拉脱'', ``电望'' and ``远镜''.}
\label{fig:examples1}
\end{figure}

\begin{figure}[t]
\centering
\begin{tabular}{|rl|}
\hline
gold standard:& 水污染 (water pollution)\\
result:& 水~~污染\\
\hline
gold standard:& 纯~~碱 (sodium carbonate,\\
&literally, pure alkali)\\
result:& 纯碱\\
\hline
\end{tabular}
\caption{There are some errors that do not contain meaningless words. In the first case, to treat ``watter pollution'' as two words ``water'' and ``pollution'' in Chinese is also grammatical and meaningful. In the second case, ``sodium carbonate'' may also be treated as a single word.}
\label{fig:examples2}
\end{figure}

Figure \ref{fig:examples1} shows some segmentation errors that result in meaningless words; while Figure \ref{fig:examples2} shows segmentation errors without producing meaningless words. There are several specifications for S\&T. Results in Figure \ref{fig:examples2} may not be wrong according to some of the specifications; while results in Figure \ref{fig:examples1} will always be wrong. Moreover, since errors with no meaningless words are still meaningful and grammatical, errors that contain meaningless words are more serious and may have more negative impact on the downstream tasks.

In this paper, we introduce word-based features as criteria for wordhood (quality of being word) in an efficient S\&T framework to reduce meaningless words and to improve the S\&T performance.

There are at least two challenges to reduce the meaningless words for S\&T.

\begin{itemize}
\item Reliable criteria for wordhood are needed for accurate word identification. The errors that contain meaningless words are caused by the limited training data. If the word ``拉脱维亚'' do not appear in the training data, it is possible for the model to guess that this is two separated new words. Thus we need extra resources for the criteria for wordhood. In this paper, we use a general lexicon, the Chinese Wikipedia and a large-scale raw corpus to provide information for the criteria. We describe the criteria in detail in Section \ref{sec:wordhood}.

\item A fast word-based S\&T framework is needed for employing word-based features we introduce. Since the word-based S\&T methods are reported much slower than the character-based ones \cite{zhang_fast_2010}, it might become intractable to employ more features. Inspired by Jiang et al. \shortcite{jiang_word_2008} and Sun \shortcite{sun_stacked_2011}, we propose an effective and efficient S\&T framework based on word-lattice and stacked learning. We describe this framework in Section \ref{sec:framework}.
\end{itemize}

We conduct our experiments on Penn Chinese Treebank 5.0. We first show that our S\&T framework outperforms the state-of-art methods in closed test with a speed of about 100 sentences per second. Then we show that after further employing features of criteria for wordhood, the number of meaningless words has a reduction of 65.7\% comparing to the baseline. As a result, the F1 measure for segmentation is increased to 0.984.

\section{Criteria for Wordhood\label{sec:wordhood}}

\subsection{General Lexicon}
Using a general lexicon is a straightforward way to determine the wordhood. If a substring $\textbf{w}$ of a sentence can be found in a general lexicon, heuristically it is possible to be segment as a word in this sentence.

SogouW\footnote{\url{http://www.sogou.com/labs/dl/w.html}} is an available large lexicon. There are 79,019 words with their word categories in this lexicon. We will use this lexicon to generate features in order to provide a criterion for wordhood.

\begin{equation}
\textup{SogouW}(\textbf{w}) = \{\tau_i\}
\end{equation}
The function returns a set containing word categories for the string $\textbf{w}$.

Note that the word categories of this lexicon do not need to be the same with the POS tags we use in our S\&T model.

\subsection{Wikipedia}
Besides general words, words that could be incorrectly segmented are named entities and domain-specific words. We use Chinese Wikipedia\footnote{\url{http://zh.wikipedia.org/}}, an online encyclopedia, as a lexicon for these words. There are totally 1,310,114 identical entries in the Chinese Wikipedia.

Unlike the general lexicon, there is no word category information about the Wikipedia entries. We simply define a function $\textup{Wiki}(\textbf{w})$ to indicate whether a string $\textbf{w}$ is an entry in Wikipedia or not:
\begin{equation}
\textup{Wiki}(\textbf{w}) = \begin{cases}
1 & \text{$\textbf{w}$ is an entity in Wikipedia}\\
0 & \text{otherwise}
\end{cases}
\end{equation}

\subsection{Large-scale Raw Corpus}

\begin{table*}[t]
\centering

\begin{tabular}{lrrrrrrrr}
\hline
$\textbf{w}$& 拉脱维亚&拉脱&维亚&射电&望远镜&电望&远镜&纯碱\\
\hline
$\textup{freq}(\textbf{w})$&203,254 &225,623&1,078,879 &63,202 &763,728 &7,546 & 766,021 & 160,607\\
$\textup{freq}(\text{\#},\textbf{w},\text{和})$ &782&1&41& 4&466&0&43&486\\
$\textup{freq}(\text{的},\textbf{w},\text{\#})$ &156&14&174 &34&10,049&0&6&534\\
\hline
$\textup{RAV}(\textbf{w})$&23&1&5&4&26&1&1&23\\
\hline
\end{tabular}
\caption{The frequences and RAV values of some strings based on the SogouT corpus.}
\label{tab:matching}
\end{table*}

Using lexicons may still have limitation, since they can not include all the low-frequency words. We use a large-scale raw corpus of web pages to provide another criterion for wordhood.

In this paper, we use SogouT\footnote{\url{http://www.sogou.com/labs/dl/t.html}} as the large-scale raw corpus. It consists of web pages with more than 200 billion Chinese characters. We use $\textup{freq}(\textbf{w})$ to indicate the frequency of string $\textbf{w}$ in this corpus, and use $\textup{freq}(l,\textbf{w},r)$ to indicate the frequency of string $\textbf{w}$ that occurs right before character $l$ and after character $r$. The symbol ``\#'' is used as a wildcard for sentence boundaries such as punctuation marks.

Large-scale raw corpus with the advantages of wide coverage and large duplication can provide rich information for language processing, although it is not manually annotated. It plays an important role in statistical machine translation \cite{brants2007large}, open information retrieval \cite{banko2009open} and other applications.

Comparing to using lexicons, we can usually find a large number of sentences that contain the concerned string $\textbf{w}$. In some of those sentences, it could be much easier to recognize that $\textbf{w}$ is a word using its context information.

Some of the examples are shown in Table \ref{tab:matching}. The string ``拉脱维亚'' occurs about 200 thousand times in SogouT. Among these occurrences, there are $\textup{freq}(\text{\#},\text{拉脱维亚},\text{和})=782$ times that ``拉脱维亚'' is at the beginning of the sentence and followed by a function word ``和 (and)''. This context suggests that ``拉脱维亚'' could be a word (at least a syntactic unit). On the contrary, ``拉脱'' barely occurs with such context, which suggests that ``拉脱'' may not be a free word.

We use restricted accessor variety (RAV) derived from the method proposed by Zhang et al.\shortcite{zhang-EtAl:2011:IJCNLP-20112} as a criterion for wordhood. There are other methods based on the distributional information to provide criterion for Chinese wordhood \cite{feng_accessor_2004,jin_unsupervised_2006}. We use RAV since it is more effective and feasible for a large-scale corpus.

Formally, unlike the original definition \cite{zhang-EtAl:2011:IJCNLP-20112}, we define a pair of characters as the restricted pairs. We use the function $\textup{match}(\textbf{w},\langle l,r\rangle)$ to indicate whether a string and a restricted pair match:
\begin{equation}
\textup{match}(\textbf{w},\langle l,r\rangle) = \begin{cases}
1 & \frac{\textup{freq}(l,\textbf{w},r)}{\textup{freq}(\textbf{w})}\geq \epsilon\\
0 & \text{otherwise}
\end{cases}
\end{equation}

Given a set $P=\{\langle l_i,r_i\rangle\}$ of restricted pairs that can match a wide range of existent words, we can define the RAV of a string as:
\begin{equation}
\textup{RAV}(\textbf{w})=\sum_{\langle l,r\rangle \in P}
{\textup{match}(\textbf{w},\langle l,r\rangle)}
\end{equation}
The larger the RAV is, the more possible that $\textbf{w}$ could be a word.

In this paper, $\epsilon$ is set to $0.0001$, and $30$ restricted pairs are automatically selected as the restricted pair set. The last row of Table \ref{tab:matching} shows the RAV values of some strings as examples.

\section{Word Lattice Based S\&T Framework\label{sec:framework}}

\begin{figure*}[t]
\centering

\begin{tabular}{crc}
\multirow{3}*{\includegraphics[width=140px]{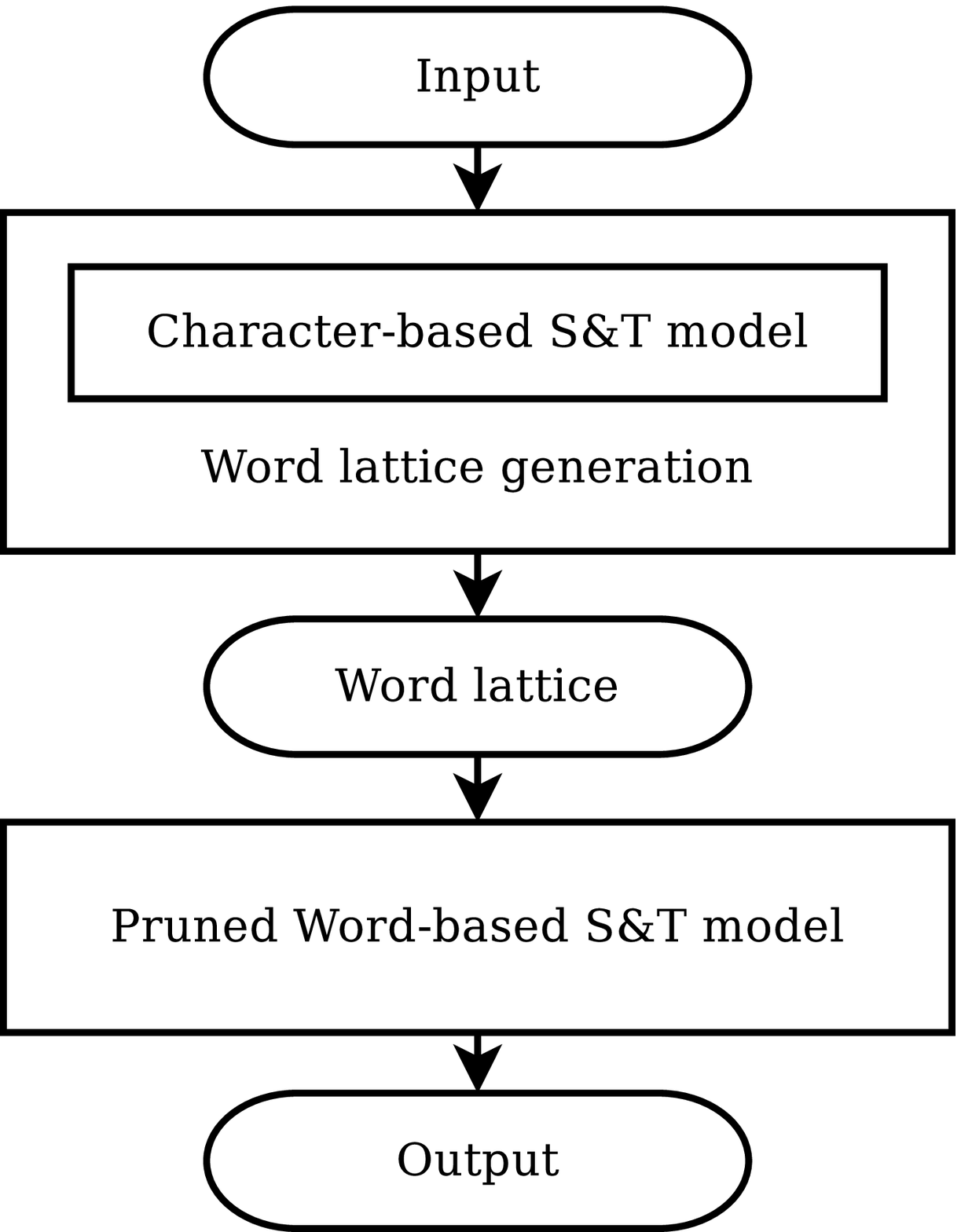}}
  &\parbox[c][40px]{40px}{Input $\textbf{c}$:}
  &\parbox[c][40px]{180px}
  {\center (……其中)在拉脱维亚驻军(最多……)} \\
&\parbox[c][120px]{70px}{Word lattice $\mathcal{L}$:}&
\parbox[c][120px]{240px}{\center\includegraphics[width=240px]{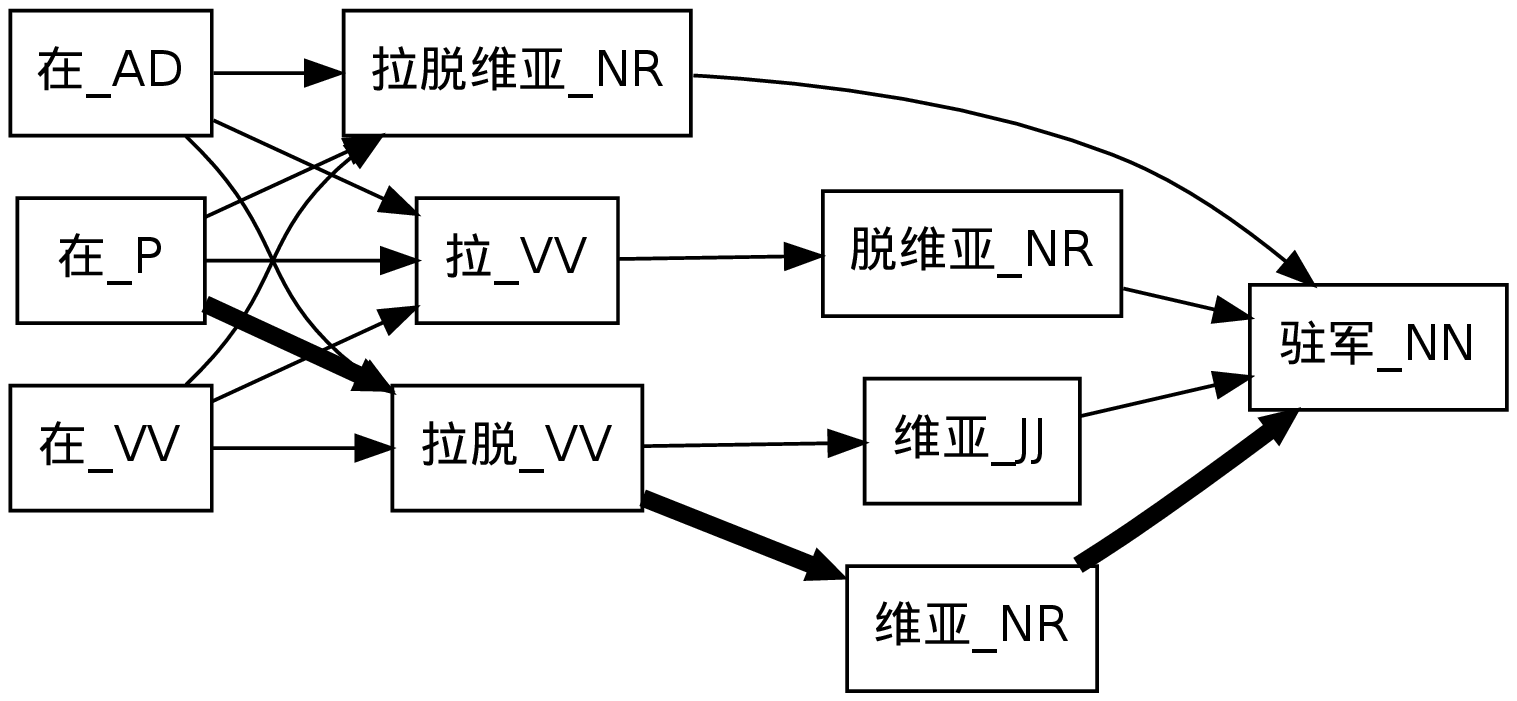}} \\
 &\parbox[c][40px]{45px}{Output $\hat{\textbf{y}}$:}
 &\parbox[c][40px]{190px}{\center\includegraphics[width=190px]{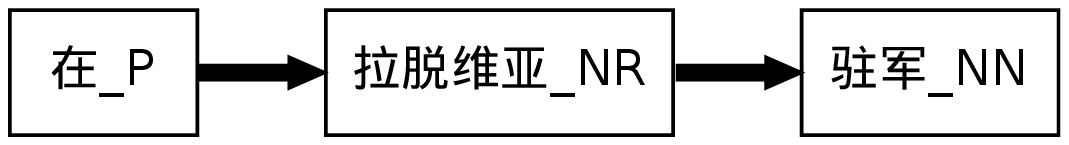}} \\
 (a)& &(b)\\
\end{tabular}

\caption{(a) shows the data flow of our S\&T system consisting of a character-based model and a pruned word-based model. There are three representations of the data, namely input $\textbf{c}$, word lattice $\mathcal{L}$ and output $\hat{\textbf{y}}$. (b) shows examples of these data in each step.}
\label{fig:workflow}
\end{figure*}

\subsection{Overview\label{subsec:overview}}
The goal of this section is to build an effective and efficient S\&T framework which can employ both character-based features and word-based features.

Figure \ref{fig:workflow} (a) is an overview of our proposed framework inspired by Jiang et al. \shortcite{jiang_word_2008} and Sun \shortcite{sun_stacked_2011}; while Figure \ref{fig:workflow} (b) shows examples of each step.

The input of S\&T is a sequence of Chinese characters:
\begin{equation}
\textbf{c}=(c_1, c_2, \dots, c_{|\textbf{c}|})
\end{equation}
where $c_i$ are Chinese characters and $|\textbf{c}|$ denotes the number of elements in the sequence $\textbf{c}$.

With the help of the character-based model (Section \ref{subsec:character_based}), a word lattice generation method (Section \ref{subsec:method_lattice}) is used to generate the word lattice which contains one or more possible outputs (Figure \ref{fig:workflow} (b)).

The word lattice is defined as a set of edges
\begin{equation}
\mathcal{L}=\{\langle p_i,\textbf{w}_i,t_i\rangle\}
\end{equation}
where each edge is a tuple $\langle p_i,\textbf{w}_i,t_i\rangle$. $\textbf{w}_i$ is the corresponding word of the edge, $t_i$ is the POS tag, and $p_i$ denotes the position which is the number of characters before $\textbf{w}_i$.

Then a pruned word-based S\&T model (Section \ref{subsec:method_word}) is used to search the final output in the word lattice. Our features for the wordhood will be employed in this model.

The output of S\&T is defined as a sequence of edges:
\begin{equation}
\textbf{y}=(\langle p_1,\textbf{w}_1,t_1 \rangle,\cdots, \langle p_{|\textbf{y}|},\textbf{w}_{|\textbf{y}|},t_{|\textbf{y}|} \rangle )
\end{equation}
where $p_i=|\textbf{w}_{1}|+\dots+|\textbf{w}_{i-1}|$.

Additionally, we use the stacked learning framework\cite{wolpert_stacked_1992} to train the character-based S\&T model and the pruned word-based S\&T model. Similar to Sun \shortcite{sun_stacked_2011}, a 10-fold cross-validation is used to train the pruned word-based model.

There is one problem that the word lattices for the training set may not always contain all the edges of the gold standard output $\textbf{y}^*$. Jiang et al.\shortcite{jiang_word_2008} provided a solution to use the ``oracle diameter'' as the gold standard in the training process of pruned word-based model. In our system, we directly add missing edges into the word lattice without defining the weights. The observation is that using our treatment is better at least in our framework.

\subsection{Character-Based Model\label{subsec:character_based}}
The character-based models for Chinese word segmentation (Seg) and S\&T are well studied (e.g., Xue and Shen \shortcite{xue:2003_c}).

The output of the character-based model is a sequence of character labels:
\begin{equation}
\textbf{a}=(\langle s_1, t_1\rangle, \dots, \langle s_{|\textbf{a}|}, t_{|\textbf{a}|}\rangle )
\end{equation}
where $|\textbf{a}|=|\textbf{c}|$. Each label is a tuple $\langle s_i, t_i\rangle$. $s_i\in S=\{\textup{b},\textup{m},\textup{e},\textup{s}\}$ is used to denote the position of $c_i$ in a word. The elements ``$\textup{b}$'' / ``$\textup{m}$'' / ``$\textup{e}$'' indicate that the character is at the beginning / in the middle / at the end a multi-character word; the element ``$\textup{s}$'' indicates that the corresponding character forms a single character word. And $t_i\in T$ is the POS tag of the word containing $c_i$.

The label $a_i=\langle s_i, t_i\rangle$ indicates the S\&T result of the corresponding character $c_i$. For example, the label $\langle \textup{b}, \textup{NN}\rangle $ indicates that the corresponding character is at the beginning of a multi-character noun with the POS tag ``NN''. These labels are used by Ng and Low \shortcite{ng_chinese_2004} and Jiang et al. \shortcite{jiang_word_2008}.

An averaged perceptron model \cite{collins_discriminative_2002} is used for the learning and prediction. Following the convention, the feature templates of our model are showed in Table \ref{tab:base_line}. The optimal label sequence $\hat{\textbf{a}}$ is calculated as:
\begin{equation}
\hat{\textbf{a}}=\argmax_{\textbf{a}}f_c(\textbf{c},\textbf{a})
\end{equation}

\begin{table}[t]
\centering

\begin{tabular}{l|l}

\hline
Unigram& $\langle c_{i-1},a_{i}\rangle$,
$\langle c_{i},a_{i}\rangle$, $\langle c_{i+1},a_{i}\rangle$ \\
Bigram& $\langle c_{i-2},c_{i-1},a_{i}\rangle$,
$\langle c_{i-1},c_{i},a_{i}\rangle$,\\
&$\langle c_{i},c_{i+1},a_{i}\rangle$,
$\langle c_{i+1},c_{i+2},a_{i}\rangle$\\
Transitional&$\langle a_{i-1},a_{i}\rangle$\\

\hline
\end{tabular}
\caption{The feature tmeplates of the character-based model}
\label{tab:base_line}
\end{table}

Roughly speaking, our model is similar to the previous proposed models with three differences. First, we do not use the feature template $\langle c_{i-1},c_{i+1},a_{i}\rangle$. According to our empirical observation, this template will cause overfitting due to the data sparseness. Second, we define special characters for sentence boundaries $c_{-1}=c_{0}=\sharp=c_{|\textbf{c}|+1}=c_{|\textbf{c}|+2}$ to generate corresponding features. Third, we find that the search order of the dynamic programming algorithm of the decoding process is important. In our model, a label is calculated earlier if it appears earlier in the gold standard of the training data.

\subsection{Word Lattice Generation\label{subsec:method_lattice}}

We will use the objective function $f_\textup{c}(\textbf{c},\textbf{a})$ of the character-based model to generate the word lattice.

The study by Jiang et al. \shortcite{jiang_word_2008} showed that using word lattice is better than using the $n$-best list for the reranking. However, in Jiang et al., the algorithm to generate the word lattice is still using a local $n$-best strategy. Here we present a more elaborate method to generate the word lattice.

In our system, the word lattice based on the input $\textbf{c}$ and a threshold $\delta$ is a set of edges, formally defined as:
\begin{equation}\label{equ:L}
\begin{split}
\mathcal{L}_{\textbf{c},\delta}=&\{\langle p_i,\textbf{w}_i,t_i\rangle | \text{there exists an output $\textbf{y}$}\\
&\text{such that } \langle p_i,\textbf{w}_i,t_i\rangle \in \textbf{\textbf{y}} \\
& \text{and } f_{\textup{c}}(\textbf{c},\hat{\textbf{a}})-
 %\left(
f_{\textup{c}}(\textbf{c},\textbf{a}(\textbf{y}))
 %\right)
 \leq \delta \}
\end{split}
\end{equation}
where $\textbf{a}(\textbf{y})$ denote the character label sequence corresponding to the output $\textbf{y}$.

The motivation is that an edge is good enough to be in the word lattice if and only if there exists an output $\textbf{y}$ containing this edge and has a high objective function value which is close to the best $f_{\textup{c}}(\textbf{c},\hat{\textbf{a}})$. Similar idea is also proposed by Huang\shortcite{huang:2008:ACLMain} and Mi et al.\shortcite{mi-huang-liu:2010:POSTERS}.

We can weight an edge $y_i=\langle p_i,\textbf{w}_i,t_i\rangle$ in the word lattice as the minimal margin between the best output $\hat{\textbf{a}}$ and the output containing $y_i$:
\begin{equation}\label{eq:weight}
m_i=f_{\textup{c}}(\textbf{c},\hat{\textbf{a}}) -
\max_{y_i\in \textbf{y}} f_{\textup{c}}(\textbf{c},\textbf{a}(\textbf{y}))
\end{equation}

\subsection{Pruned Word-based Model\label{subsec:method_word}}
Given a word lattice, our system needs to find a sequence of edges to form the output. This can be formalized as:
\begin{equation}
\hat{\textbf{y}}=\argmax_{y_i\in \mathcal{L}_{\textbf{c},\delta}}{f_{\textup{w}}(\textbf{y})}
\end{equation}
We call this pruned word-based model since we only consider words from $\mathcal{L}_{\textbf{c},\delta}$.

%Note that $\mathcal{L}_{\textbf{c},\infty}=\textup{Cand}_{\textbf{c}}$. The architecture of the straightforward word-based model \cite{zhang_joint_2008} can be seen as a special case of ours.

Also, an averaged perceptron model is used for the learning and prediction. Word-based features based on the criteria in Section \ref{sec:wordhood} are used in this model.

\begin{table}[t]
\centering

%<tag',l2',tag,l2> <tag',tag>

\begin{tabular}{l|l}
\hline

1 & $\langle \textbf{w}_i \rangle$ \\
2 & $\langle \textbf{w}_i, t_i \rangle$ \\
3 & $\langle |\textbf{w}_i|>1 \rangle$  \\
4 & $\langle h(m_i) \rangle$  \\
5 & $\langle h(m_i), |\textbf{w}_i|>1 \rangle$  \\
6 & $\langle t_i \rangle$ \\
7 & $\langle t_i, |\textbf{w}_i|>1 \rangle$ \\
8 & $\langle t_{i-1}, t_i \rangle$ \\
9 & $\langle t_{i-1},|\textbf{w}_{i-1}|>1,t_i, |\textbf{w}_i|>1 \rangle$ \\
\hline
10 & $\langle t_i, \tau\in\textup{SogouW}(\textbf{w}_i)\rangle$ \\
11 & $\langle t_i, \textup{Wiki}(\textbf{w}_i)\rangle$ \\
12 & $\langle t_i, \lceil\textup{RAV}(\textbf{w}_i)/2\rceil\rangle$ \\
\hline
\end{tabular}
\caption{The feature templates of the pruned word-based S\&T model. Templates 10 to 12 are only for the open test.}
\label{tab:templates_w}
\end{table}

The feature templates used for the pruned word-based model are listed in Table \ref{tab:templates_w}. Templates 1 to 9 are used for both closed and open test; while Templates 10 to 12 related to the wordhood are only used for the open test.

Template 1 represents the word corresponding to the edge. Since word length is useful, we use Template 3 to represent such information. Template 8 and Template 9 are used to represent the information of two contiguous words.

Stacked features (related to the previous character-based model) are also represented as Template 4 and Template 5. The weight of an edge $m_i$ is defined in Eq. \ref{eq:weight}. $h(m)$ is a genralized step function that maps a continuous value to a discrete value: $h(m)=\lceil\log_2{\lceil m \rceil}\rceil$ and we define $h(0)=-\infty$.

Template 10 is based on the SogouW. If a word is assigned with more than one word categories in SogouW, each of the categories $\tau$ will generate a feature together with $t_i$. The machine learning algorithm will automatically learn the relation between the word categories and the POS tags. Template 11 is based on the Wikipedia. And Template 12 is based on the RAV calculated from the large-scale corpus. For the limitation of time and storage, we do not calculate and store the RAV for strings with frequency less than 1,000.

\section{Experiments\label{sec:exp}}
We conduct our experiments on the Penn Chinese Treebank 5.0 (CTB 5.0). We use the same training, development and test sets as the previous work \cite{jiang_word_2008,zhang_fast_2010,sun_stacked_2011}. Table \ref{tab:CTB5_sets} shows the summary of these sets. The development set is used to determine some parameters of our system such as the feature templates and the threshold $\delta$ for the word lattice generation.

\begin{table}[t]
\centering

\begin{tabular}{lrrr}
\hline
Data set & CTB files & Sentences & Words\\

\hline

Training & 1-270&18,086&493,938\\
&400-931\\
&1001-1151\\
Dev.&301-325&350&6,821\\
Test&271-300&348&8,008\\

\hline
\end{tabular}
\caption{Summary of the training, development and test sets of CTB 5.0}
\label{tab:CTB5_sets}
\end{table}

An edge is correct if and only if it appears in both $\hat{\textbf{y}}$ and the gold standard output sequence $\textbf{y}^*$. We use the F1 measure to measure the performance of our S\&T system. the F1 for S\&T is defined as:
\begin{equation}
\textup{F1}=\frac{2\sum_\textbf{c}{|\hat{\textbf{y}} \cap \textbf{y}^*|}}
{\sum_\textbf{c}{|\hat{\textbf{y}}|}+\sum_\textbf{c}{|\textbf{y}^*|}}
\end{equation}
where $|\hat{\textbf{y}} \cap \textbf{y}^*|$ is the number of correct edges of sentence $\textbf{c}$.

If we do not distinguish POS tags of edges in $\hat{\textbf{y}}$ and $\textbf{y}^*$ in the equation, we get the definition of the F1 for Chinese word segmentation (Seg). We use this F1 measure to evaluate the performance of Seg.

\subsection{Influence of Lattice Scale}

Word lattices are generated with the help of the character-based model and the word lattice generation algorithm described in Section \ref{subsec:method_lattice}. The parameter $\delta$ is used to control the lattice scale.

We define the {\it lattice scale} which is the ratio of the size of the word lattice to the size of the gold standard:
$\textup{lattice scale}={\sum_\textbf{c}{|\mathcal{L}_{\textbf{c},\delta}|}}/
{\sum_\textbf{c}{|\textbf{y}^*|}}$. And we define the {\it oracle recall} which is the ratio of the number of the edges in both the word lattice and the gold standard to the size of the gold standard to indicate in what degree the gold standard outputs are contained in the word lattices: $
\textup{oracle recall}={\sum_\textbf{c}{|\mathcal{L}_{\textbf{c},\delta} \cap \textbf{y}^*|}}
/{\sum_\textbf{c}{|\textbf{y}^*|}}
$. The oracle recall is similar to the oracle F1 used by related work to indicate the quality of the first step of a two-step S\&T method \cite{jiang_word_2008,sun_stacked_2011}.

\begin{figure}[t]
\centering

\includegraphics[width=200px]{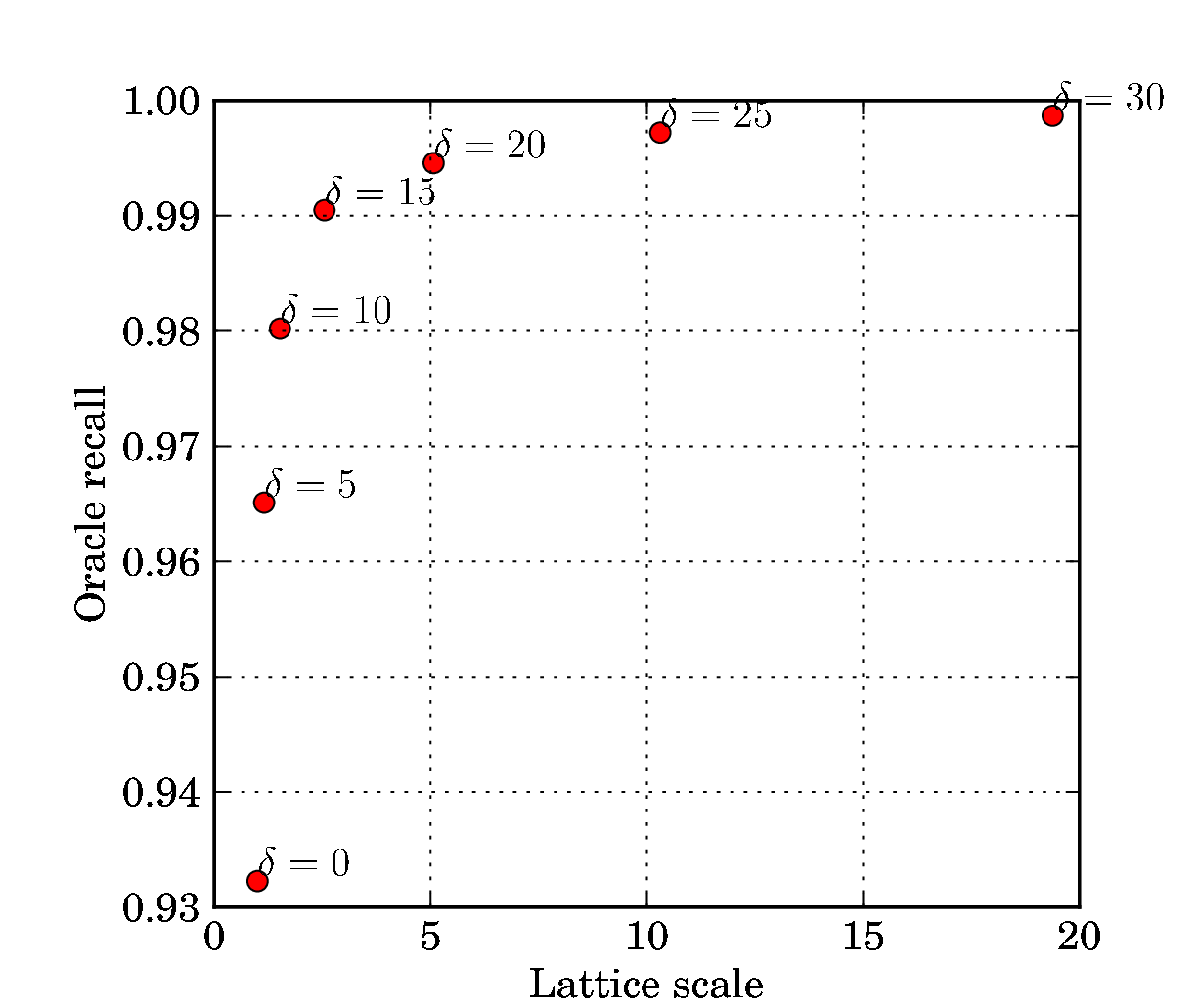}\\

\caption{The lattice scale and the oracle recall can be easily manipulated by changing the threshold $\delta$}
\label{fig:oracle}
\end{figure}

In our system, we can easily manipulate the lattice scale and the oracle recall in a wide range by changing $\delta$. The relation between there three variables are shown in Fig. \ref{fig:oracle}. %For example, when $\delta=30$, the oracle recalls for Seg and S\&T are $0.9999$ and $0.9987$, respectively.

As shown in \cite{jiang_word_2008}, when the degree is set to $10$, the lattice scale for the test set is larger than $10$ and the oracle F1 is $0.9779$. The oracle F1 reported by Sun \shortcite{sun_stacked_2011} is 0.9929, which relies on three models.

Then, we find a local optimal $\delta$ to determine the lattice scale used for the pruned word-based model. Results on the development set are shown in Table \ref{tab:lattice_delta}. Feature Templates 1 to 9 are used. We set $\delta$ to $15$ in the rest of our experiments. Note that a large $\delta$ will also cause a longer decoding time.

\begin{table}[t]
\centering
\begin{tabular}{lrr}
\hline
$\delta$ & Seg & S\&T\\
\hline
13 & 0.9676&0.9339\\
14 & 0.9678&0.9336\\
15 & {\bf 0.9700} & {\bf 0.9354}\\
16 & 0.9692&0.9341\\
17 & 0.9685&0.9337\\

\hline
\end{tabular}
\caption{The influence of the threshold $\delta$ on the development set}
\label{tab:lattice_delta}
\end{table}

\subsection{Closed Test for Pruned Word-based Model}

We conduct the experiment on the test set for closed test and compare our results with the previous work in Table \ref{tab:closed_result}.

Our system outperforms the previous systems listed in the table in closed test according to the Seg F1 and the S\&T F1. Especially, our system has a high performance of Seg F1.

\begin{table}[t]
\centering
\begin{tabular}{lrr}
\hline
Closed test & Seg & S\&T\\
\hline
\cite{jiang_cascaded_2008}&0.9785&0.9341\\
\cite{jiang_word_2008}&0.9774&0.9337\\
\cite{kruengkrai_error-driven_2009}&0.9787&0.9367\\
\cite{zhang_fast_2010}&0.9778&0.9367\\

\hline
character-based model&0.9790&0.9336\\
full system&\textbf{0.9811}&\textbf{0.9383}\\
~~95\% confidence interval&$\pm$0.0030&$\pm$0.0054\\
\hline
\end{tabular}
\caption{Comparison of our model and related work for the closed test.}
\label{tab:closed_result}
\end{table}

We perform the significant test using method for Chinese word segmentation in SIGHAN bakeoff \cite{emerson_second_2005}. The 95\% confidence intervals of our full system are listed below the F1 measures. We see that the differences between those methods are not significant enough. This issue is for all the related work.

We measure the speed of our system for the closed test on the test set. The program uses one core of 2.53GHz (Intel Core 2 Duo E7200). Results are shown in Table \ref{tab:speed}. The total time of the test procedure is 3.229 seconds, including 1.041 seconds for lattice generation and 2.188 seconds for the pruned word-based model. And note that the feature extraction of the pruned word-based model (a pruned word-based S\&T model) is implemented using Python, which is much slower than C/C++. The speed of the S\&T model using cross validation is reported by Zhang and Clark \shortcite{zhang_fast_2010}. The speed of that word-based S\&T model is 24.92 sentences per second (a single 2.66GHz Intel Core 2 CPU), while the speed of our system is 107.77 sentences per second.

\begin{table}[t]
\centering
\begin{tabular}{lr}
\hline
Total time &3.229 sec.\\
\quad Lattice generation&1.041 sec.\\
\quad Pruned word-based model&2.188 sec.\\
Speed&107.77 sent./sec.\\
\hline
\end{tabular}
\caption{The speed of our system on the test set consisting of 348 sentences for the closed test}
\label{tab:speed}
\end{table}

\subsection{Features for Wordhood}

Finally, we conduct the experiment on the test set for open test and compare our results with the previous work in Table \ref{tab:open_result}.

We try different combinations of our three feature templates for wordhood. We find that in both development set and test set, features derived from the general lexicon and the raw corpus are helpful while features derived from the Wikipedia is not helpful.

The 95\% confidence intervals are calculated for the best results of our model. The differences are also not significant.

\begin{table}[t]
\centering
\begin{tabular}{lrr}

\hline
Open test& Seg & S\&T\\
\hline
\cite{jiang_automatic_2009}&0.9823&0.9403\\
\cite{sun_stacked_2011}&0.9817&0.9402\\
\cite{wang-EtAl:2011:IJCNLP-2011}&0.9811&{\bf 0.9417}\\

\hline
w/ SogouW &0.9834&0.9403\\
w/ SogouT &0.9828&0.9399\\
w/ SogouW, SogouT &{\bf 0.9844}&0.9411\\
~~95\% confidence interval&$\pm$0.0028&$\pm$0.0053\\
w/ SogouW, Wiki, SogouT  &0.9841&0.9404\\
\hline
\end{tabular}
\caption{Comparison of our model and related work for the open test.}
\label{tab:open_result}
\end{table}

\subsection{Analysis of Meaningless Words}
We use the method proposed by Li and Sun \shortcite{li_punctuation_2009} to manually check every errors in the result. We count the number of errors that are not caused by granularity and results in meaningless words.

\begin{figure}[t]
\centering

\includegraphics[width=200px]{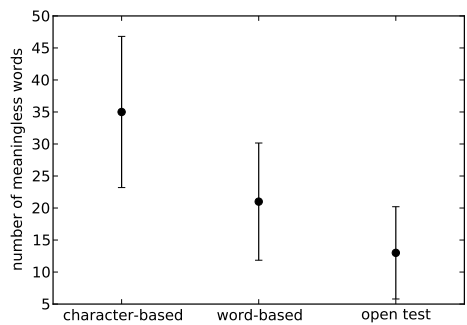}\\

\caption{The numbers of meaningless words with 95\% confidence intervals of our method.}
\label{fig:error}
\end{figure}

The numbers of meaningless words with 95\% confidence intervals are shown in Figure \ref{fig:error}. Since we can not get the corresponding results for the related work, we only count the results of our character-based model, pruned word-based model and the model employing features of wordhood.

In comparison with the character-based model, the number of meaningless words of our system with features of wordhood has a significant reduction. The number of meaningless words has a reduction of 62.9\%, while the number of total incorrect words has a reduction of 25\%.

\section{Related Work\label{sec:related_work}}
Zhang and Clark \shortcite{zhang_joint_2008}, Mochihashi et al. \shortcite{mochihashi_bayesian_2009} and Kruengkrai et al. \shortcite{kruengkrai_error-driven_2009} proposed models for S\&T or Seg with both character-based features and word-based features in a single model of which the time spending is considerable.

Sun \shortcite{sun_stacked_2011} proposed a stacked framework to connect a character-based model and a word-based model in parallel; while Jiang et al. \shortcite{jiang_word_2008} proposed a word lattice based reranking method in which the character-based model and word-based model are connected in series. The main difference between the method by Jiang et al. and ours are the lattice generation method and the feature templates. And we use the way proposed by Sun to train the character-based model and word-based model.

Several resources were used for the open test of S\&T and Seg. Sun\shortcite{sun_stacked_2011} used a lexicon of idioms. Jiang et al. \shortcite{jiang_automatic_2009} used an annotated corpus with different annotation standard. Wang et al. \shortcite{wang-EtAl:2011:IJCNLP-2011} used auto-analyzed data. The data used in those methods are not comparable with the corpus of ours.

Li and Sun \shortcite{li_punctuation_2009} used nearly the large-scale corpus with ours in a semi-supervised way. The learned model had a better ability to recognize new word, while it did not outperform the baseline. In this paper, we employ the statistical information for that corpus as features in our model.

There are proposed criteria for Chinese word extraction \cite{chien_pat-tree-based_1997,chen_unknown_2002}. For example, Feng et al.\shortcite{feng_unsupervised_2004,feng_accessor_2004} proposed the criterion accessor variety for Chinese word extraction and used it for Chinese word segmentation. Jin et al.\shortcite{jin_unsupervised_2006} used branching entropy to perform unsupervised Chinese word segmentation. Since these methods will become intractable for a large scale corpus, we adapted the method by Zhang et al.\shortcite{zhang-EtAl:2011:IJCNLP-20112} in our experiment.

\section{Conclusion\label{sec:conclusion}}

In this paper, we proposed an S\&T method based on the word lattice adapted from Jiang et al.\shortcite{jiang_word_2008}. A simple but more effective method is used for the word lattice generation based on the objective function of a character-based S\&T model. The scale of lattice can be easily controlled by a simple parameter $\delta$. Then a pruned word-based S\&T model is used to find an optimal path in the word lattice as the output. Our experiments on Penn Chinese treebank 5 show that our system is faster than the single word-based model and outperforms the previous work in the closed test.

We employ features of Chinese wordhood from lexicon and large-scale raw corpus of 200 billion characters. Our final system has a 62.9\% reduction of meaningless word generation in comparison with the character-based model. As a result, the F1 measure for segmentation is increased to 0.984.

The CTB corpus mainly consists of news and magazines. And the performance is already high. When we want to do S\&T on documents such as blogs, microblogs or domain-specified documents other than news, the meaningless words in the output will be an even greater challenge. We will test our method on such documents in the future.
\bibliographystyle{acl2012}
\bibliography{bib}

\begin{thebibliography}{}

\bibitem[\protect\citename{Banko \bgroup et al.\egroup }2009]{banko2009open}
M.~Banko, M.J. Cafarella, S.~Soderland, M.~Broadhead, and O.~Etzioni.
\newblock 2009.
\newblock {\em Open information extraction for the web}.
\newblock University of Washington.

\bibitem[\protect\citename{Brants \bgroup et al.\egroup }2007]{brants2007large}
T.~Brants, A.C. Popat, P.~Xu, F.J. Och, and J.~Dean.
\newblock 2007.
\newblock Large language models in machine translation.
\newblock In {\em In EMNLP}. Citeseer.

\bibitem[\protect\citename{Chen and Ma}2002]{chen_unknown_2002}
{Keh-Jiann} Chen and {Wei-Yun} Ma.
\newblock 2002.
\newblock Unknown word extraction for chinese documents.
\newblock In {\em Proceedings of the 19th international conference on
  Computational {linguistics-Volume} 1}, pages 1--7. Association for
  Computational Linguistics.

\bibitem[\protect\citename{Chien}1997]{chien_pat-tree-based_1997}
{Lee-Feng} Chien.
\newblock 1997.
\newblock {PAT-tree-based} keyword extraction for chinese information
  retrieval.
\newblock In {\em {ACM} {SIGIR} Forum}, volume~31, pages 50--58. {ACM}.

\bibitem[\protect\citename{Collins}2002]{collins_discriminative_2002}
Michael Collins.
\newblock 2002.
\newblock Discriminative training methods for hidden markov models: Theory and
  experiments with perceptron algorithms.
\newblock In {\em Proceedings of the {ACL-02} conference on Empirical methods
  in natural language {processing-Volume} 10}, page 1每8.

\bibitem[\protect\citename{Emerson}2005]{emerson_second_2005}
Thomas Emerson.
\newblock 2005.
\newblock The second international chinese word segmentation bakeoff.
\newblock In {\em Proceedings of the Fourth {SIGHAN} Workshop on Chinese
  Language Processing}, pages 123--133. Jeju Island, Korea.

\bibitem[\protect\citename{Feng \bgroup et al.\egroup
  }2004a]{feng_accessor_2004}
Haodi Feng, Kang Chen, Xiaotie Deng, and Weimin Zheng.
\newblock 2004a.
\newblock Accessor variety criteria for chinese word extraction.
\newblock {\em Computational Linguistics}, 30(1):75--93.

\bibitem[\protect\citename{Feng \bgroup et al.\egroup
  }2004b]{feng_unsupervised_2004}
Haodi Feng, Kang Chen, Chunyu Kit, and Xiaotie Deng.
\newblock 2004b.
\newblock Unsupervised segmentation of chinese corpus using accessor variety.
\newblock {\em Natural Language Processing {IJCNLP} 2004}, pages 694--703.

\bibitem[\protect\citename{Huang}2008]{huang:2008:ACLMain}
Liang Huang.
\newblock 2008.
\newblock Forest reranking: Discriminative parsing with non-local features.
\newblock In {\em Proceedings of ACL-08: HLT}, pages 586--594, Columbus, Ohio,
  June. Association for Computational Linguistics.

\bibitem[\protect\citename{Jiang \bgroup et al.\egroup
  }2008a]{jiang_cascaded_2008}
Wenbin Jiang, Liang Huang, Qun Liu, and Yajuan L邦.
\newblock 2008a.
\newblock A cascaded linear model for joint chinese word segmentation and
  part-of-speech tagging.
\newblock In {\em In Proceedings of the 46th Annual Meeting of the Association
  for Computational Linguistics}.

\bibitem[\protect\citename{Jiang \bgroup et al.\egroup }2008b]{jiang_word_2008}
Wenbin Jiang, Haitao Mi, and Qun Liu.
\newblock 2008b.
\newblock Word lattice reranking for chinese word segmentation and
  {Part-of-Speech} tagging.
\newblock In {\em Proceedings of the 22nd International Conference on
  Computational {Linguistics-Volume} 1}, pages 385--392. Association for
  Computational Linguistics.

\bibitem[\protect\citename{Jiang \bgroup et al.\egroup
  }2009]{jiang_automatic_2009}
Wenbin Jiang, Liang Huang, and Qun Liu.
\newblock 2009.
\newblock Automatic adaptation of annotation standards: Chinese word
  segmentation and {POS} tagging 牢 a case study.
\newblock In {\em Proceedings of the 47th {ACL}}, page 522每530, Suntec,
  Singapore, August. Association for Computational Linguistics.

\bibitem[\protect\citename{Jin and {Tanaka-Ishii}}2006]{jin_unsupervised_2006}
Zhihui Jin and Kumiko {Tanaka-Ishii}.
\newblock 2006.
\newblock Unsupervised segmentation of chinese text by use of branching
  entropy.
\newblock In {\em Proceedings of the {COLING/ACL} 2006 Main Conference Poster
  Sessions}, page 428每435, Sydney, Australia, July. Association for
  Computational Linguistics.

\bibitem[\protect\citename{Kruengkrai \bgroup et al.\egroup
  }2009]{kruengkrai_error-driven_2009}
Canasai Kruengkrai, Kiyotaka Uchimoto, Jun'ichi Kazama, Yiou Wang, Kentaro
  Torisawa, and Hitoshi Isahara.
\newblock 2009.
\newblock An {Error-Driven} {Word-Character} hybrid model for joint chinese
  word segmentation and {POS} tagging.
\newblock In {\em Proc. of {ACL-IJCNLP} 2009}, page 513嚙瘠521, Suntec,
  Singapore. Association for Computational Linguistics.

\bibitem[\protect\citename{Li and Sun}2009]{li_punctuation_2009}
Zhongguo Li and Maosong Sun.
\newblock 2009.
\newblock Punctuation as implicit annotations for chinese word segmentation.
\newblock {\em Computational Linguistics}, 35(4):505--512.

\bibitem[\protect\citename{Mi \bgroup et al.\egroup
  }2010]{mi-huang-liu:2010:POSTERS}
Haitao Mi, Liang Huang, and Qun Liu.
\newblock 2010.
\newblock Machine translation with lattices and forests.
\newblock In {\em Coling 2010: Posters}, pages 837--845, Beijing, China,
  August. Coling 2010 Organizing Committee.

\bibitem[\protect\citename{Mochihashi \bgroup et al.\egroup
  }2009]{mochihashi_bayesian_2009}
Daichi Mochihashi, Takeshi Yamada, and Naonori Ueda.
\newblock 2009.
\newblock Bayesian unsupervised word segmentation with nested {Pitman-Yor}
  language modeling.
\newblock In {\em Proceedings of the Joint Conference of the 47th Annual
  Meeting of the {ACL} and the 4th International Joint Conference on Natural
  Language Processing of the {AFNLP}}, page 100每108.

\bibitem[\protect\citename{Ng and Low}2004]{ng_chinese_2004}
Hwee~Tou Ng and Jin~Kiat Low.
\newblock 2004.
\newblock Chinese part-of-speech tagging: one-at-a-time or all-at-once?
  word-based or character-based.
\newblock In {\em Proc of {EMNLP}}.

\bibitem[\protect\citename{Sun}2011]{sun_stacked_2011}
Weiwei Sun.
\newblock 2011.
\newblock A stacked {Sub-Word} model for joint chinese word segmentation and
  {Part-of-Speech} tagging.
\newblock In {\em Proceedings of the 49th Annual Meeting of the Association for
  Computational Linguistics: Human Language Technologies}, page 1385嚙瘠1394,
  Portland, Oregon, {USA}, June. Association for Computational Linguistics.

\bibitem[\protect\citename{Wang \bgroup et al.\egroup
  }2011]{wang-EtAl:2011:IJCNLP-2011}
Yiou Wang, Jun'ichi Kazama, Yoshimasa Tsuruoka, Wenliang Chen, Yujie Zhang, and
  Kentaro Torisawa.
\newblock 2011.
\newblock Improving chinese word segmentation and pos tagging with
  semi-supervised methods using large auto-analyzed data.
\newblock In {\em Proceedings of 5th International Joint Conference on Natural
  Language Processing}, pages 309--317, Chiang Mai, Thailand, November. Asian
  Federation of Natural Language Processing.

\bibitem[\protect\citename{Wolpert}1992]{wolpert_stacked_1992}
David Wolpert.
\newblock 1992.
\newblock Stacked generalization*.
\newblock {\em Neural networks}, 5(2):241--259.

\bibitem[\protect\citename{Xue and Shen}2003]{xue:2003_c}
Nianwen Xue and Libin Shen.
\newblock 2003.
\newblock Chinese word segmentation as {LMR} tagging.
\newblock In {\em Proceedings of the second {SIGHAN} workshop on Chinese
  language {processing-Volume} 17}, pages 176--179. Association for
  Computational Linguistics.

\bibitem[\protect\citename{Zhang and Clark}2008]{zhang_joint_2008}
Yue Zhang and Stephen Clark.
\newblock 2008.
\newblock Joint word segmentation and {POS} tagging using a single perceptron.
\newblock In {\em Proceedings of {ACL}}, volume~8.

\bibitem[\protect\citename{Zhang and Clark}2010]{zhang_fast_2010}
Yue Zhang and Stephen Clark.
\newblock 2010.
\newblock A fast decoder for joint word segmentation and {POS-Tagging} using a
  single discriminative model.
\newblock In {\em {EMNLP-10}}.

\bibitem[\protect\citename{Zhang \bgroup et al.\egroup
  }2011]{zhang-EtAl:2011:IJCNLP-20112}
Kaixu Zhang, Ruining Wang, Ping Xue, and Maosong Sun.
\newblock 2011.
\newblock Extract chinese unknown words from a large-scale corpus using
  morphological and distributional evidences.
\newblock In {\em Proceedings of 5th International Joint Conference on Natural
  Language Processing}, pages 837--845, Chiang Mai, Thailand, November. Asian
  Federation of Natural Language Processing.

\end{thebibliography}

\end{CJK}
\end{document}